\documentclass{article}

\usepackage{PRIMEarxiv}

\usepackage{graphicx}      
\usepackage{amsfonts}
\usepackage{subcaption}
\usepackage{amssymb}
\usepackage{amsmath}
\usepackage{natbib}        
\usepackage{booktabs}
\usepackage{wrapfig}
\usepackage{float}
\usepackage{array}
\usepackage[nolist]{acronym} 

\usepackage{pifont} 

\newcommand{\cmark}{\ding{51}} 
\newcommand{\xmark}{\ding{55}} 

\usepackage{xcolor} 

\usepackage{placeins}
\usepackage{makecell}

\begin{document}

\title{Evaluating Modern Visual Anomaly Detection Approaches in Semiconductor Manufacturing: A Comparative Study}

\author{
Manuel Barusco \\
University of Padova, Italy \\
\texttt{manuel.baruscophd.unipd.it} \\ \And
Francesco Borsatti \\
University of Padova, Italy \\
\texttt{francesco.borsatti.1@phd.unipd.it} \\ \And
Youssef Ben Khalifa \\
University of Padova, Italy \\
\texttt{youssef.benkhalifa@studenti.unipd.it} \\ \And
Davide Dalle Pezze \\
University of Padova, Italy \\
\texttt{davide.dallepezze@unipd.it} \\ \And
Gian Antonio Susto \\
University of Padova, Italy \\
\texttt{gianantonio.susto@unipd.it} \\
}

\maketitle

\begin{abstract}                
Semiconductor manufacturing is a complex, multistage process.
Automated visual inspection of Scanning Electron Microscope (SEM) images is indispensable for minimizing equipment downtime and containing costs.
Most previous research considers supervised approaches, assuming a sufficient number of anomalously labeled samples.
On the contrary, Visual Anomaly Detection (VAD), an emerging research domain, focuses on unsupervised learning,  avoiding the costly defect collection phase while providing explanations of the predictions.
We introduce a benchmark for VAD in the semiconductor domain by leveraging the MIIC dataset.
Our results demonstrate the efficacy of modern VAD approaches in this field.
\end{abstract}

\keywords{Visual Anomaly Detection \and Computer Vision \and Semiconductor Manufacturing}


\section{Introduction}
\label{sec:introduction}

Semiconductor manufacturing is a highly complex, multistage process that transforms raw silicon into functional integrated circuits (ICs).
In the semiconductor sector, to ensure high product quality and production performance, defect detection is a relevant problem that can reduce scrap rates, minimize equipment downtime, and cut overall manufacturing costs.
Therefore, automated visual inspection, particularly via Scanning Electron Microscope (SEM) imaging of wafers, has become a critical component of the manufacturing workflow \cite{qiao2025ra,luo2025scsnet}.

The advent of deep learning has further advanced defect detection, delivering impressive results \cite{qiao2024deepsem,luo2025scsnet,chen2024towards}.
Most of the literature considers the supervised paradigm, where labeled images from both normal and abnormal classes are present in the training set.
However, obtaining and annotating sufficient anomalous SEM images is often impractical.
In the real world, this is an unrealistic assumption given the difficulty in obtaining and annotating sufficient anomalous images.

Visual Anomaly Detection (VAD) addresses this challenge by learning only from defect-free samples.
Moreover, they produce anomaly maps that highlight the pixels responsible for abnormality scores.
This makes the model explainable in its predictions, allowing human operators to understand not just that an anomaly has occurred, but where and potentially why.
This aspect is vital in the Industry 5.0 framework, which emphasizes the human-centric paradigm, where AI systems must not only perform accurately but also facilitate human oversight and support collaborative decision-making.

To date, only a single study has investigated VAD in the context of semiconductor manufacturing \cite{huang2021}. That work introduced an SEM‐based wafer dataset and evaluated three reconstruction‐based algorithms. However, since that early effort, the field has shifted dramatically: nowadays, most of the leading methods are feature‐based, offering superior accuracy and robustness compared to their reconstruction‐based predecessors.

For this reason, our first contribution is to provide a benchmark in Visual Anomaly Detection for the semiconductor manufacturing field.
In particular, we introduced several state-of-the-art approaches, including 7 approaches that were never tested in this domain.

Moreover, from a broader perspective of the VAD field, ours is one of the few studies that examine the behavior of VAD methods on SEM images.
Indeed, although VAD has gained much attention over the past few years, the research has targeted natural-image domains, such as the MVTec dataset \cite{MVTec}, with objects such as screws, hazelnuts, etc.
In contrast, specialized imaging modalities like SEM remain underexplored. 

The main contributions of our work can be summarized as
follows:
\begin{itemize}
    \item We provide a benchmark for automated visual inspection in semiconductor manufacturing. 
    \item We prove the effectiveness of modern feature-based VAD approaches in localizing the defects of the SEM images.
    \item We provide a detailed analysis and discussion of the SEM domain compared to natural images.
\end{itemize}

The outline of the paper is as follows.
In Section 2, we review related work in defect detection for the semiconductor field.
Section 3 details our methodology: first, the MIIC dataset and its relevance in the semiconductor field and, more generally, in VAD research are described.
Then, more details about VAD are provided: the nine VAD methods and the VAD metrics considered in this work are listed.
In Section 4, the experimental setting is described, including details like hyperparameters used for the tested VAD methods.
In Section 5, we present quantitative results, comparing image-level and pixel-level performance across all methods, and analyze their behavior on SEM data. 
We conclude in Section 6 by summarizing our contributions and outlining directions for future research.

\begin{figure}[h]
    \centering
    \includegraphics[width=0.48\textwidth]{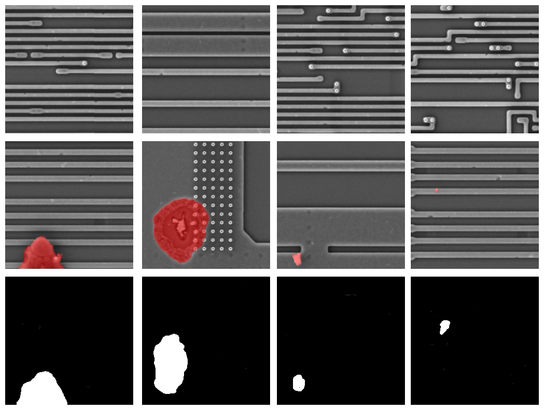}
    \caption{Representative images from the MIIC dataset. In the first row are reported normal images, in the second row anomalous images with the anomalies highlighted in red, in the last row the segmentation masks produced by CFA.} 
    \label{fig:MIIC_dataset}
\end{figure}

\section{Related Work} 
\label{sec:related_work}

\subsection{Semiconductor Manufacturing}
\label{subsec:related_work_semiconductor}

Semiconductor manufacturing is a highly intricate, multistep process that transforms raw silicon wafers into functional integrated circuits.
The semiconductor fabrication includes several stages, such as oxidation, thin-film deposition, photolithography, etching, ion implantation, and chemical mechanical planarization.
These six steps are iterated, often dozens of times, to build up the device’s multilayer structures. 
After completing all patterning layers, the flow continues with metallization, back-end processing, packaging, testing, and inspection.

Automated Visual Inspection tools applied to SEM images can detect defects, which is crucial to reduce waste and production costs while maximizing yield.
SEM images are high-resolution, grayscale visualizations of the microscopic surface morphology of a sample.
The goal is to identify defects that can be represented as particles, scratches, cracks, or pattern deformations that are otherwise invisible to optical microscopes.

For this task, data-driven techniques proved their usefulness.
For example, \cite{chen2024towards} formulates defect inspection as a supervised object detection problem.
They build on the YOLO-NAS backbones and add a Super-Resolution branch to learn high-resolution features.
Instead, in \cite{qiao2024deepsem}, authors propose to integrate Convolutional Neural Networks (CNNs) with Transformers to enhance defect classification and precise segmentation.
Specifically, they consider a classification branch and a segmentation branch.

\cite{qiao2025ra} adopts a U-shaped architecture that leverages residual networks and a novel attention module to enhance the ability of the network to focus on defects while suppressing background noise.
Moreover, they address severe class imbalance through data augmentation and a specific loss function.

Eventually, \cite{luo2025scsnet} proposes a method that is a fusion of CNN and Transformer and uses Multi-Cross-Attention Fusion (MCF) modules, plus residual simplification in a U-Net-style decoder.

While these methods prove the usefulness of Deep Learning applied to SEM images of semiconductor manufacturing processes, they still rely on supervised datasets.
However, acquiring a sufficient set of anomalous samples is extremely challenging in real-world scenarios, given the rarity of these samples and considering that the pixel-level annotations can be a time-consuming and resource-intensive process.
Moreover, the SEM datasets used in these studies are not publicly released, which hinders reproducibility and prevents fair, direct comparison of new techniques.

In contrast, authors in \cite{huang2021} consider the defect detection task in an unsupervised manner and also make available the dataset called MIIC.
However, since they performed this study at the early stages of the VAD field, the current SOTA approaches are missing.
In particular, they tested only reconstruction-based approaches, while current SOTA approaches belong mainly to feature-based approaches.
Moreover, while the image-level performance is provided, the pixel-level performance is missing.

Therefore, we benchmark state-of-the-art VAD methods and report performance using both image-level and pixel-level metrics.
In addition, we provide an analysis of the behavior of modern VAD approaches when applied to images that diverge from natural domains, which is not well investigated.

\begin{table*}[!thb]
\centering
\caption{Comparison of different anomaly detection datasets. Good = number of normal images present in the dataset; Bad = number of anomalous images present in the dataset. In bold, the dataset used in our study. Note that it is the only large-scale dataset with SEM images that deviate from the classic natural domain.}
\label{tab:comparison_datasets}
\begin{tabular}{@{}lcccccccc@{}}
\toprule
Dataset             & Authors                 & Year & Classes & Images & Domain    & Defects    & Good & Bad \\
\midrule
NanoTwice           & \cite{carrera2016defect}    & 2016 & 1       & 45     & SEM        & real-world & 40   & 5   \\
MVTEC AD            & \cite{MVTec}      & 2019 & 15      & 5354   & natural    & simulated  & 4096  & 1258   \\
KSSD2               & \cite{KSSD2}        & 2021 & 1       & 3335   & natural    & real-world & 2979 & 356 \\
BTAD                & \cite{mishra2021vt}         & 2021 & 3       & 2540   & natural    & real-world & 2250  & 290  \\
\textbf{MIIC}       & \cite{huang2021}         & 2021 & 1       & 25276   & SEM        & real-world & 25160& 116   \\
VisA                & \cite{zou2022spot}         & 2022 & 12      & 10821  & natural    & simulated  & 9,621  & 1200   \\
\bottomrule
\end{tabular}
\end{table*}

\section{Methodology}
\label{sec:methodology}

\subsection{Benchmark}
\label{subsec:benchmark}

MIIC (Microscopic Images of Integrated Circuits) is a large-scale dataset consisting of real high-resolution microscopic images of integrated circuits \cite{huang2021}.
It is the first publicly available, large-scale dataset specifically designed for the task of Visual Anomaly Detection in the semiconductor domain.
All images are captured at the metal layer of fully manufactured ICs.

The dataset includes a total of 25,276 grayscale SEM (Scanning Electron Microscopy) images, of which 25,160 are normal and 116 contain various types of anomalies. 
Each image has a resolution of 512 $\times$ 512 pixels, offering fine-grained visual detail necessary for inspection.
Fig. \ref{fig:MIIC_dataset} presents some samples of normal and anomalous images, showing also the pixel-wise ground truth mask for the abnormal images.

In Tab. \ref{tab:comparison_datasets}, we compare VAD datasets in terms of dataset size, number of classes, and publication year.
The table highlights that most publicly available VAD benchmarks for industrial inspection are drawn from natural-image domains. 
By contrast, SEM microscopy captures images within an entirely distinct visual domain.
It produces high-resolution, single-channel scans with characteristic noise patterns, intricate surface textures, and defect morphologies that are fundamentally distinct from those found in photographic datasets.

Excluding the MIIC dataset, the only other dataset containing SEM images is the NanoTwice dataset \cite{carrera2016defect}.
Although both NanoTwice and MIIC fall under the SEM modality and both contain real-world structural anomalies, they are produced from entirely different manufacturing processes.
Indeed, NanoTwice is from nanofibrous materials, while MIIC is from integrated-circuit wafers, and thus they exhibit distinct defect characteristics. 
Moreover, the two datasets differ dramatically in scale: MIIC offers over 25000 defect-free images, facilitating the training of modern deep models, whereas NanoTwice has a limited size of only 45 images.
In addition, it should be stressed that the MIIC dataset presents a high level of heterogeneity, as suggested by examples shown in Fig. \ref{fig:MIIC_dataset}, implying that the variety among images is extremely high, making understanding the difference between normal and abnormal challenging.

Before the release of MVTec AD \cite{MVTec} in 2019, NanoTwice was among the few resources for visual anomaly detection. 
The debut of MVTec AD, with over 5000 images spanning 15 object and texture categories, rapidly established it as the field’s standard benchmark. 
As a result, subsequent research has overwhelmingly focused on natural-image datasets like MVTec AD and Visa.
On these benchmarks, excellent results were obtained with feature-based approaches.
These methods exploit a feature extractor to obtain meaningful image patch representations from the small target dataset.
However, these models are usually pretrained on datasets like ImageNet, which consists almost entirely of natural photographs, representing a significant domain gap when these features are applied to SEM microscopy images.
Therefore, it is difficult to assess a priori that the state-of-the-art feature-based VAD approaches would work well in the semiconductor field, since it is not certain that the feature extractor can produce meaningful representations for this domain.
We discuss this aspect in detail in Sec. \ref{subsec:discussion}.

\subsection{Visual Anomaly Detection}

VAD methods aim to identify samples that deviate from the normal behavior seen during training.
In particular, its goal is to detect in an unsupervised manner abnormal images and locate anomalous regions of the image.
In this way, it is possible to avoid the time-consuming and resource-intensive labeling process.
Moreover, VAD improves explainability and user confidence by providing an anomaly map of which pixels are considered anomalous and caused the entire image to be classified as abnormal.

Nowadays, VAD methods fall into two categories: \\reconstruction-based and embedding-based methods. \\
\textbf{Reconstruction-based methods} use generative models to learn normal image reconstruction during training: large reconstruction errors during inference indicate anomalies. Approaches basedo on generative models such as AutoEncoders, Generative Adversarial Networks (GANs) are prominent in this area.

Specifically, \cite{huang2021} tested the following two reconstruction-based approaches: GANomaly and f-anoGAN.
f-AnoGAN uses a two-step approach.
Initially, it trains a WGAN to learn the distribution of the normal samples.
Then it trains an encoder to map images to the latent space.

In contrast to f-anoGAN, GANomaly trains all components simultaneously.
GANomaly is composed of a conditional GAN and an additional encoder, with the generator of GAN structured as an encoder-decoder architecture.
During training, the model learns the distribution of normal samples by minimizing the distance between the original and reconstructed latent vectors while also minimizing the reconstruction loss between the original and reconstructed images.
At inference, anomalies are detected by the difference between the embedding produced by the encoder of the generator and the embedding produced by the additional encoder.

Another method, called RIAD, uses the inpainting technique to detect anomalies \cite{RIAD_inpaint}.
The idea from RIAD is to randomly remove small regions from anomaly-free images and train a model to inpaint them. 
At inference, it masks multiple patches, and poorly inpainted ones are considered anomalous.

The work of \cite{huang2021} also proposed a novel method called IAD+Inpainting.
They use a similar architecture to GANomaly, though they apply some modifications like sharing the weight for the encoder of the generator $G_E$ and the second encoder $E$.
In a first training phase, the model learns to reconstruct the images.
Then, in the second phase, they apply the inpainting technique.

\textbf{Feature Embedding-based methods} rely on data representations generated by pre-trained neural networks.
These approaches are particularly appealing not only because of their high performance but also because if the chosen backbone is lightweight, then these methods can be executed efficiently, even on resource-constrained or edge devices \cite{barusco2024paste}.

These approaches can be further categorized into three categories as follows.

\textbf{Teacher-Student approaches} based on two networks (teacher and student) with knowledge distillation, where student and teacher feature map deviations indicate anomalies, with the most famous approach of this category being STFPM \cite{st_pyramid}. Subsequently, RD4AD \cite{rd4ad} improved the STFPM approach by considering an autoencoder-like approach where the teacher is the encoder and the student the decoder.
\\
\textbf{Memory Bank approaches} save features of normal images in a memory bank. Approaches like Padim, PatchCore, and CFA fall under this category.
Padim models each spatial feature location with a multivariate Gaussian and uses the Mahalanobis distance to detect deviations as anomalies \cite{PaDiM}. 
PatchCore builds a compact memory bank of representative normal patches and flags anomalies based on the distance of test patches to their nearest neighbors \cite{patch}.
Then CFA creates a memory of normal patch embeddings and adapts them into coupled hyperspheres to amplify the separation between normal and abnormal feature representations \cite{lee2022cfa}.
Eventually, SuperSimpleNet also uses a feature adaptor like CFA, but it also considers the generation of synthetic anomalies at feature-level to improve performance  \cite{rolih2025supersimplenet}.
\\
\textbf{Normalizing Flow approaches} based on normalizing flow models to transform complex input data distributions into normal distributions, leveraging probability as a measure of normality.
A notable example in this family is FastFlow \cite{yu2021fastflow}.

\subsection{Metrics}

The performance of Visual Anomaly Detection techniques is typically evaluated using a range of metrics. To provide a comprehensive assessment and a fair comparison with the original work, we employ pixel and image-level F1, ROC and PR scores. 
In addition to these metrics, we also utilize the Per-region-overlap (PRO) metric for pixel-level evaluation. The PRO metric ensures that ground-truth regions are weighted equally, regardless of their size, thereby mitigating the limitations of simplistic per-pixel metrics \cite{yang2020improving}.

\begin{table}[!thb]
\centering
\begin{tabular}{|c|c|c|c|c|}
\hline
\textbf{VAD Method} & \textbf{Year} & \textbf{Type} & \makecell{\textbf{Previous}\\\textbf{Study}} & \textbf{Our} \\ \hline
\textbf{GANomaly}         & 2018 &      R          & \cmark & \cmark \\ \hline
\textbf{f-anoGAN}           & 2019 &      R          & \cmark & \cmark \\ \hline
\textbf{IAD+Inpainting}       & 2021 &      R          & \cmark & \cmark \\ \hline
\textbf{PaDiM}            & 2021 &      F          & \xmark & \cmark \\ \hline
\textbf{STFPM}            & 2021 &      F          & \xmark & \cmark \\ \hline
\textbf{CFA}              & 2022 &      F          & \xmark & \cmark \\ \hline
\textbf{PatchCore}        & 2022 &      F          & \xmark & \cmark \\ \hline
\textbf{RD4AD}             & 2022 &      F          & \xmark & \cmark \\ \hline
\textbf{FastFlow}         & 2022 &      F          & \xmark & \cmark \\ \hline
\textbf{SuperSimpleNet}   & 2025 &      F          & \xmark & \cmark \\ \hline
\end{tabular}
\caption{Comparison of Visual Anomaly Detection (VAD) methods: year, type, inclusion in previous studies, and in our evaluation. Type can be R: \textit{reconstruction-based} or F: \textit{features-based}. We focus on the second family of approaches since current SOTA approaches belong to this family. }
\label{tab:VAD_approaches}
\end{table}

\subsection{Experimental Setup}
\label{subsec:experimental_setup}

\begin{table*}[!t]
\centering
\begin{tabular}{|c|ccc|cccc|}
\hline
\textbf{VAD Method}      & \multicolumn{3}{c|}{\textbf{Image‐level (\%)}} & \multicolumn{4}{c|}{\textbf{Pixel‐level (\%)}}                      \\ \hline
                         & ROC     & F1      & PR      & ROC     & F1      & PR      & PRO     \\ \hline
\textbf{f-anoGAN*}          & 56.38   & 31.65   & –       & –       & –       & –       & –       \\ \hline
\textbf{GANanomaly*}      & 93.34   & 74.64   & –       & –       & –       & –       & –       \\ \hline
\textbf{IAD+Inpainting*}      & 99.27   & 91.23   & –       & –       & –       & –       & –       \\ \hline
\textbf{PatchCore}       & 93.33   & 75.51   & 78.22   & 98.64   & 57.67   & 44.88   & 93.03   \\ \hline
\textbf{PaDiM}           & 84.89   & 47.55   & 47.29   & 99.05   & 22.98   & 13.86   & 95.30   \\ \hline
\textbf{CFA}             & 99.01   & 87.96   & 94.28   & 99.96   & 75.66   & 78.91   & 99.04   \\ \hline
\textbf{STFPM}           & 96.32   & 77.83   & 83.31   & 99.97   & 77.30   & 80.77   & 99.18   \\ \hline
\textbf{R4AD}            & 93.93   & 68.48   & 72.28   & 99.85   & 56.94   & 54.92   & 98.45   \\ \hline
\textbf{FastFlow}        & 97.63   & 78.70   & 86.17   & 99.60   & 53.02   & 46.03   & 99.57   \\ \hline
\textbf{SuperSimpleNet}  & 96.23   & 80.72   & 86.65   & 98.46   & 50.57   & 43.84   & 97.59   \\ \hline
\end{tabular}
\caption{Detection (image‐level) and localization (pixel‐level) performance of the evaluated VAD methods on the MIIC dataset, expressed as percentages. Methods with * report the results from the original paper; otherwise, the method is implemented and evaluated with our code.}
\label{tab:results}
\end{table*}

For the training, the dataset was split following the original train-test division, keeping 23888 normal images for the training set, while using 116 abnormal images and 1274 normal images for the test set. Half of the test set was considered as a validation set for saving the best model during training.
All images are rescaled to size 224x224 and normalized with the mean and the standard deviation of the ImageNet dataset, as is usually performed in VAD methods.

All evaluations were conducted on an Ubuntu-based workstation equipped with an NVIDIA L40S GPU, an AMD AMD EPYC 9224 24-Core Processor CPU, and 1511 GB of system RAM. The python version is 3.8 and CUDA version is 12.2.
We used all the hyperparameters originally used in the methods when tested on MVTec in their original work. 
Also, the feature extractor and the feature extraction layers are those considered in the original works.
For the PatchCore model, we limited training to a random subset comprising 50\% of the available images to mitigate the otherwise prohibitive memory requirements for performing the coreset reduction needed for the memory bank construction.

\begin{figure*}[!h]
  \centering
  \begin{subfigure}[b]{0.45\textwidth}
    \centering
    \includegraphics[width=\linewidth]{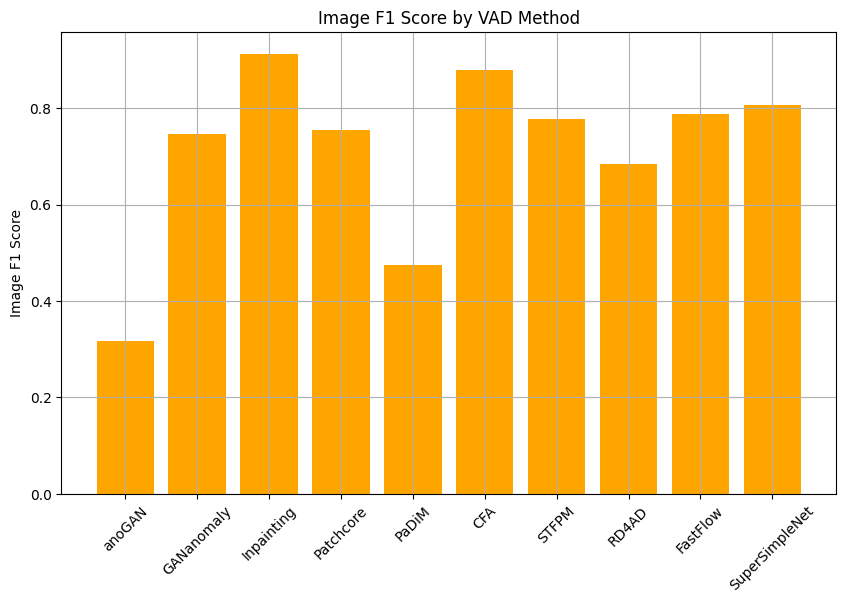}
    \caption{Image F1 Score}
    \label{fig:f1-img}
  \end{subfigure}
  \hfill
  \begin{subfigure}[b]{0.45\textwidth}
    \centering
    \includegraphics[width=\linewidth]{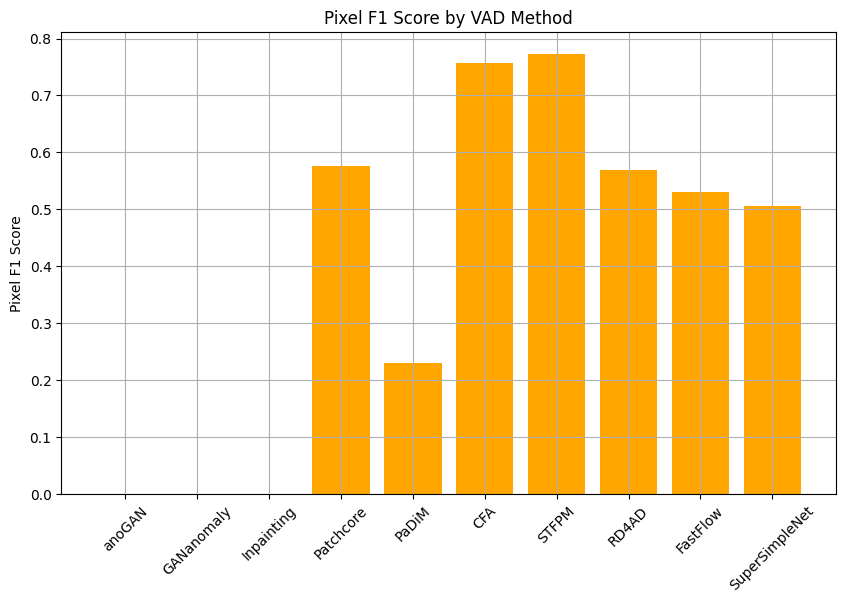}
    \caption{Pixel F1 Score}
    \label{fig:f1-pxl}
  \end{subfigure}

  \caption{Comparison of F1 scores by VAD method on image level(left) and on pixel level(right).}
  \label{fig:f1-comparison}
\end{figure*}

\section{Results}
\label{sec:results}

\subsection{Feature-based Approaches}

We first analyze the results for each method for the image-level performance, reported in Tab. \ref{tab:results} and illustrated also in Fig. \ref{fig:f1-img}, which represents the ability to detect an image as anomalous.
At the image level, most feature-based approaches yield comparable ROC-AUC scores on MIIC: PatchCore achieves 93.3\%, FastFlow 97.6\%, STFPM 96.3\%, and CFA 99.0\%.
Notably, PaDiM (84.89\%) underperforms other models despite reporting good results in the MVTec; furthermore, STFPM, which performs worse than RD4AD in the MVTec dataset, is working better here.
This discrepancy highlights the importance of domain-specific evaluation: on SEM data (MIIC), good architectures tested on natural images do not necessarily translate to better detection tools.

Turning to the image-level F1 metric, distinctions among feature-based methods become more pronounced (presented in Tab. \ref{tab:results} and Fig. \ref{fig:f1-pxl}), yet CFA again leads (87.9\%), followed by FastFlow (78.7\%) and STFPM (77.8\%). 
This superior performance of CFA is likely due to its auxiliary concentration network, which more effectively clusters normal feature representations.

When considering the pixel-level metrics (F1 and PRO), the model with the highest F1 score is the STFPM, followed by CFA. 
This highlights that methods that are the best in defect detection (image level), are not necessarily the best for defect localization (pixel level).
By looking at the PRO metric, those differences are not evident, with all the methods achieving very high values. Nevertheless, the F1 score for anomaly detection tasks is better because it takes into consideration the intrinsic problem of unbalanced datasets.

\subsection{Discussion}
\label{subsec:discussion}

In this section, we compare the results obtained from feature-embedding approaches with classical reconstruction-based techniques.
Based on this comparison, we discuss what these findings imply for the broader VAD research, particularly the need to diversify the benchmark beyond natural photographs and the need for large-scale datasets.

Surprisingly, the inpainting-based reconstruction model, despite being a pretty old technique, still rivals the feature-based methods, with 99.3 \% ROC and 91.2 \% F1 at the image level, which can be compared only to the CFA method.

We believe this could be due to different reasons:
(i) the ability to produce meaningful representations of a feature extractor pre-trained on ImageNet (photographs), which differ significantly from the target domain (SEM images).
(ii) Contrary to most VAD datasets that have hundreds of images for a class, the MIIC is a large-scale dataset containing more than 25000 images.
This abundance of data makes it easier for a reconstruction-based model to learn well the underlying patterns.

These results suggest two key takeaways for future VAD research:
\begin{itemize}
    \item \textbf{Domain diversity matters}. Benchmarks should extend beyond photographic datasets (e.g., MVTec, VisA) to include specialized modalities like SEM, where pretrained feature extractors may be less optimal.

    \item \textbf{Revisiting reconstruction}. Given sufficient normal data, modern reconstruction-based methods may able to match or surpass feature embeddings. 
\end{itemize}

By broadening the scope of VAD evaluation to cover both novel domains and large-scale collections, we can better understand when and why each methodological paradigm excels.

\section{Conclusion}
\label{sec:conclusion}

In this work, we have introduced a benchmark of modern VAD approaches for the semiconductor manufacturing field, leveraging the MIIC dataset.
We evaluate this dataset with six feature-based and three reconstruction-based approaches using both image-level and pixel-level metrics.
This benchmark can be used for both researchers and practitioners to improve the performance in approaches developed for the semiconductor field and in general when dealing with images that deviate from natural photographs.

Our analysis proved that VAD approaches are highly explainable, and they align closely with the human-centric goals of Industry 5.0, helping the user in the decision-making process.
Moreover, despite the significant domain gap between natural photographs and electron-microscope data, our analysis demonstrates that pretrained feature extractors remain a robust foundation for defect detection in the semiconductor field.
This means that feature-based VAD approaches remain a valid choice even when dealing with different domains.

As possible future research directions, based on our analysis, it would be interesting to pay more attention to reconstruction-based VAD approaches.
They remain unaffected by the target domain, and given the availability of a large-scale training dataset, these models should capture well the structure of the normal images.
Moreover, feature-based methods could be improved by considering feature extractors fine-tuned to the considered dataset instead of being only pretrained on ImageNet. CFA is an example of how a feature refinement step could benefit the final performance. The fine-tuning could be done by considering self-supervised techniques that do not require labels (not available in the classic VAD scenario). This could lead to more specific features that can help the feature-based methods.

\bibliographystyle{unsrt}  
\bibliography{references}

\end{document}